\pdfoutput=1

\documentclass[11pt]{article}

\usepackage[]{acl}

\usepackage{times}
\usepackage{latexsym}

\usepackage[T1]{fontenc}

\usepackage[utf8]{inputenc}

\usepackage{microtype}
\usepackage{multirow}

\usepackage{graphicx}

\usepackage{soul}
\newcommand{\cut}[1]{}

\newcommand{\rs}[1]{\textcolor{red}{RS: #1}}

\newcommand{\example}[1]{\textit{``#1''}}
\newcommand{\gloss}[1]{\textit{[#1]}}

\newcommand{\cdc}{CDC}
\newcommand{\dsname}{CoRAL}
\newcommand{\dslongname}{\textbf{Co}ntext-aware \textbf{C}roatian \textbf{A}busive \textbf{L}anguage Dataset}

\title{CoRAL: a Context-aware Croatian Abusive
Language Dataset}

\author{Ravi Shekhar$^{1}$, Mladen Karan$^{1}$, Matthew Purver$^{1,2}$ \\
  $^{1}$Cognitive Science Research Group,  School of Electronic Engineering \& Computer Science,\\
  Queen Mary University of London, UK \\
  $^{2}$Department of Knowledge Technologies,
  Jožef Stefan Institute, Ljubljana, Slovenia\\
  \texttt{\{r.shekhar, m.karan, m.purver\}@qmul.ac.uk} \\}

\date{}

\begin{document}
\maketitle
\begin{abstract}
In light of unprecedented increases in the popularity of the internet and social media, comment moderation has never been a more relevant task. Semi-automated comment moderation systems greatly aid human moderators by either automatically classifying the examples or  allowing the moderators to prioritize which comments to consider first. However, the concept of inappropriate content is often subjective, and such content can be conveyed in many subtle and indirect ways. In this work, we propose \dsname{}\footnote{ The \dsname{} dataset can be found \href{https://github.com/shekharRavi/CoRAL-dataset-Findings-of-the-ACL-AACL-IJCNLP-2022}{here}.} -- a language and culturally aware Croatian Abusive dataset covering phenomena of implicitness and reliance on local and global context. We show experimentally that current models degrade when comments are not explicit and further degrade when language skill and context knowledge are required to interpret the comment.

\end{abstract}

\section{Introduction}
\label{sec:Intro}

The growing volume of user-generated content -- from social media to online forums and comments under news articles -- implies a growing need for moderation of this content to counter abuse and the spread of misinformation. Automatic and semi-automatic moderation systems 
can greatly aid human moderators, making their work quicker, easier, and more accurate; however, most of this work focuses on English, ignoring smaller, less-resourced languages~\citep{vidgen2020directions}. This situation is improving with the advent of multilingual contextual language models, as they enable cross-lingual transfer learning: recent work shows that comment moderation models with reasonable performance for less-resourced languages can be produced using zero- or few-shot transfer learning after pre-training on majority language datasets~\citep{pelicon-etal-2021-zero, pelicon2021investigating}.

It is not always sufficient to identify whether a comment is inappropriate or not; further sub-categorization helps build measures to counter it. Previous work has taken a range of approaches to sub-categorizing inappropriate content. \citet{waseem2017understanding} divided abusive language into two orthogonal categories -- directed/generalized and implicit/explicit. A very similar approach is taken by \citet{zampieri2019predicting}. 
More fine-grained approaches include very specific topics such as \textit{homophobia}, \textit{cyberbullying} or \textit{racism} \citep[e.g.,][]{Mollas2022ETHOSAM}, and the annotation of community-specific extreme hate speech with targets from multiple countries \citep{maronikolakis-etal-2022-listening}; we refer to \citet{poletto2021resources} for a comprehensive list. Recently, a unified taxonomy of abusive language categories has been proposed by \citet{banko2020unified}, a systematic division of slurs by \citet{kurrek2020towards}, and another taxonomy by \citet{Fortuna2019AHP}.  \citet{Rttger2021HateCheckFT,rottger-etal-2022-multilingual} provide a detailed empirical analysis of model performance across different example categories. All of these approaches divide comments primarily on the basis of how/whom they insult. In contrast, we are interested in categorizing how such comments can be difficult to classify or interpret automatically due to their use of linguistic and cultural context.

Our goal is to create a dataset and accompanying annotation schema to quantify what categories (primarily related to linguistic and cultural context) of abuse are being used by people and how well NLP models handle these different categories. To this end, we identified three context dependency categories (\cdc): Implicitness, Global Context, and Local Context. These \cdc s are further sub-divided according to implicitness (explicit/implicit), use of (global/local) language alterations, and use of (global/local) external knowledge; see Section~\ref{sec:dataset} for details. The closest related work in this vein is that of \citet{wiegand2021implicitly}, who give a systematic overview of various ways in which examples can be difficult (e.g., sarcasm, dehumanization, inference required, multimodality, etc.). However, \citet{wiegand2021implicitly} only focused on  implicit abuse in English without any empirical analysis.

We focus on the Croatian language, a less-represented language in Natural Language Processing research. We annotated 2,240 Croatian comments from the 24sata newspaper\footnote{\url{https://www.24sata.hr/}} with our proposed \cdc s. We experimented with four transformer-based models~\citep{mbert, ulcar-robnik2020finest,ljubesic-lauc-2021-bertic,conneau2020unsupervised}. 
Our experimentation shows that models do not perform equally well on all \cdc s. 
The easiest \cdc{} is explicit expression (e.g., cursing or using slurs), confirming the findings of \citet{wiegand2019detection}. More difficult \cdc s are those that require global or local context for their interpretation, via language disguise or external knowledge.

The contribution of this paper is twofold. First, we present a publicly available schema and the \emph{\dslongname} (\dsname) comprised of Croatian news comments annotated for different \cdc s. 

Second, we provide a quantitative and qualitative comparison of comment moderation models, revealing the limitations of different cross-lingual models when handling difficult examples and which \cdc s are generally the most challenging.

\section{Dataset}
\label{sec:dataset}

\begin{table*}[ht]
\resizebox{\textwidth}{!}{
\centering
\small
\begin{tabular}{l|r|r|r|r||r|r||r}
\hline
 &  \multicolumn{4}{c||}{\# Vote} & \multicolumn{2}{c||}{Majority Votes}& \multirow{2}{*}{$\kappa$} \\
\cline{2-4} \cline{5-7}    
  & 0 & 1 & 2 & 3 & w Expl.& w/out Expl. &   \\ \hline 
Explicit  Expression & 506  & 425   & 484 & 825 & 1,309  &  - & 0.45  \\ 
Implicit Expression & 1,297  &  567 & 275   &  101 & 376 & 363  & 0.25   \\ \hline
Language Independent Disguise & 1,941  & 95   & 78  & 126 & 204  & 99  & 0.70    \\
World Knowledge-Based & 1,136  & 571  & 357   &  176 & 533 &  163 & 0.31    \\ \hline
Croatian Specific Disguise & 1,642  & 312  & 193  & 93 &  286 & 146   & 0.30   \\
Croatia Knowledge-Based & 2,155   &  55  & 26  & 4 &  30 &  14  & 0.40  \\ \hline

Others & 1,866  & 198 & 103& 73  & 176 & 175 & 0.47 \\
\hline
Total & - & - & - & - & 2,240 & 931 & -  \\\hline
\end{tabular}
}
 \caption{Dataset Statistics: First, we report the number of annotators voted(0-3) for \cdc s. Then we report with/without Explicit Expression \cdc{} and inter-annotator agreement (Fleiss' $\kappa$), based on the majority votes(i.e., 2 or 3 votes) . The  “w/out Explicit” columns for all cases when it is not labeled as Explicit.}
\label{table:dataSpecificity}
\end{table*}

\begin{table*}[ht]
\resizebox{\textwidth}{!}{
\centering
\small
\begin{tabular}{l|r|r|r|r}

  & \# disagreements & sample size & \# ambiguous  & \# majority ok \\ \hline 
Explicit  Expression & 909 (40.6\%) & 91 & 70 (76.9\%) & 79 (86.8\%) \\
Implicit Expression & 842 (37.6\%) & 85 & 62 (72.9\%) & 64 (75.3\%) \\ \hline
Language Independent Disguise & 173 ~~(7.7\%) & 50 & 36 (72.0\%) & 41 (82.0\%) \\
World Knowledge-Based & 81 ~~(2.6\%) & 50 & 47 (92.2\%) & 43 (84.3\%) \\ \hline
Croatian Specific Disguise & 928 (41.4\%) & 93 & 70 (75.3\%) & 73 (78.5\%) \\
Croatia Knowledge-Based & 505 (22.5\%) & 51 & 47 (92.1\%) & 38 (74.5\%) \\ \hline

Others & 301 (13.4\%) & 50 & 40 (76.9\%) & 43 (82.7\%) \\
\hline
\end{tabular}
}
 \caption{Analysis of data ambiguity. Columns are (1) number of examples with disagreement for a \cdc, (2) size of the sample we annotated, (3) number of examples from the sample annotated as ambiguous (4) number of examples from the sample where the fourth annotator agrees with the majority \cdc{} label of the remaining three.}
\label{table:dataAmbiguity}
\end{table*}

When building \dsname, we aim to have annotated examples with the \cdc's they exhibit. Moreover, we focus on devising \cdc s that would reflect the challenges models face when accounting for cultural context (global or local). By manual inspection, we identified three main \cdc s of blocked comments on which cross-lingual models tend to fail: 
\textit{Implicitness}, \textit{Global context}, and \textit{Local-context}, which are further divided as follows.

\begin{itemize}

    \item \textbf{Implicitness}: Defines whether examples express abuse  directly or indirectly.  
    
        \begin{itemize}
            \item \textbf{Explicit Expression}: 
            directly use abusive words, e.g., 
            derogation, threatening language, 
            slurs, profanity. (e.g.\ \example{Retardiran si.} \gloss{You are retarded.})
        
            \item \textbf{Implicit Expression}:
            use indirect ways to express abuse, usually through vague statements implying abuse without stating it, e.g., sarcastic compliments. (e.g.\ \example{Pametan si ko panj.} \gloss{You're as smart as a stump.})
        \end{itemize}
    
    \item \textbf{Global Context}:  Defines if general background knowledge  is required.
       
        \begin{itemize}
            \item \textbf{Language Independent Disguise}: Linguistic alterations applicable in any language. E.g.,  adjacent character swap, missing characters/word boundaries, extra spaces, etc. (e.g.\ \example{J**i se.} \gloss{F**k you.})
            
            \item \textbf{World Knowledge-Based}: 
            The comment requires world/global knowledge (e.g., globally known characters, events, or facts) to be fully understood.(e.g.\ \example{Adolf je bio u pravu.} \gloss{Adolf was right.})

        \end{itemize}

    \item \textbf{Local Context}: Defines if Croatia-specific background knowledge  is required. 
    
    \begin{itemize}
        \item \textbf{Croatian Specific Disguise}: Linguistic alterations 
           specific to the Croatian language. E.g., ad-hoc constructed words that are understandable to locals, missing/wrong diacritics, using dialects, etc. (e.g.\ \example{Promijenit ću ti lični opis.} \gloss{I will change your personal description. -- I will break your face. })

        \item \textbf{Croatia Knowledge-Based}: 
         The comment requires Croatia specific knowledge (e.g., local characters, events, or facts) to be fully understood. (e.g.\ \example{Treba tebe u Vrapče.} \gloss{You need to be put into Vrapče -- Vrapče is a famous mental asylum in Croatia.)}

    \end{itemize}
    \item \textbf{Other}: Anything else not covered above

\end{itemize}

To the best of our knowledge, \dsname{} is the first dataset with annotations on which category of local/global context is  
required for interpretation.
\footnote{See Appendix 1 for examples of each \cdc.}

\paragraph{Dataset Annotation:} We use the publicly available 24sata newspaper comment dataset~\citep{shekhar-etal-2020-automating}.\footnote{Available at \url{https://clarin.si/repository/xmlui/handle/11356/1399}~\citep{pollak-etal-2021-embeddia}} The dataset contains comments moderated by 24sata's moderators based on the newspaper's policy: rules include the removal of hate speech, abusive statements, threats, obscenity, deception
\& trolling, vulgarity, and comments that are not in Croatian. We refer to  \citet{shekhar-etal-2020-automating} for more details, reproduced here in Appendix 2.

We randomly selected \cut{\hl{Matt: is this a new random split? Or the same as Pelicon/Shekhar? In which case we shouldn't say ``we'' did it} \rs{new test}} 2,240 blocked comments from 2019 related to abuse only (i.e., 24sata's abuse, hate speech, obscenity, and vulgarity categories).  
We take a \emph{multi-label} approach: annotators were 
asked to select all (possibly multiple) \cdc s they think apply to the comment; if none 
applies, then select \textit{Other} and provide an explanation. 
 
Three annotators annotated each comment from a total of 7 annotators we had available. All annotators are university students and paid on an hourly basis. Each annotator was provided training and feedback during three pilots. 

\paragraph{Dataset Statistics:}  In Table~\ref{table:dataSpecificity}, we present the statistics of the dataset based on the majority \cdc{} label.
More than 58\% of blocked comments is from \textit{Explicit Expression} \cdc, followed by \textit{Croatian Specific Disguise} (23\%). To further gain insight into the data, we remove all comments marked \textit{Explicit Expression} \cdc. In that case, most comments were from the \textit{Implicit Expression} \cdc, followed by \textit{Croatian Specific Disguise}. The \textit{World Knowledge-Based} comments were less than 1.5\%, which might be due to a small volume of world-related articles on the 24sata newspaper. 

\paragraph{Inter-Annotator Agreement:} The inter-annotator agreement, measured by Fleiss' $\kappa$ \cite{fleiss1971measuring} is moderate or better ($\geq 0.4$) for 4/7 \cdc s and fair ($\geq 0.2$) for the rest (see Table~\ref{table:dataSpecificity}). We get the lowest agreement on the \textit{Implicit Expression} \cdc{} (0.25),\cut{\hl{Matt: be more specific, give actual kappa figures}} likely due to this \cdc{} being very subjective. On the other hand, the best agreement is on \textit{Language Independent Disguise} (0.70), which is the most clearly defined \cdc.

To further explore agreement, we divided the data into 4 subsets for every \cdc, based on the number of annotators who gave a positive vote.  0 and 3 therefore correspond to perfect agreement between the three annotators, while 1 and 2 are disagreement. In Table~\ref{table:dataSpecificity}, we provide the statistics of this division. To gain additional insight into the structure of disagreements we sampled 10\% (but no fewer than 50) of examples with disagreement for each
\cdc{} ( 
see Table~\ref{table:dataAmbiguity}). One of the authors then annotated these examples with a fourth ``expert'' \cdc{} label. 
This additional label matched  
the majority label in more than 75\% of cases for each \cdc{} label (Table~\ref{table:dataAmbiguity}, majority column). This indicates that many disagreements could be resolved by additional annotation or use of majority voting; but also that many examples with disagreement are genuinely ambiguous with no clear-cut obviously ``correct'' choice  for the \cdc{} label (multiple choices were all valid to an extent). Consequently, we opted not to force  resolution of disagreements, but rather to leave them as part of the data.\footnote{We release all individual annotations, not only the majority vote based decisions.} We next explore this ambiguity in more detail.

Some tasks are inherently subjective/ambiguous, and their disagreements can never be completely resolved --- see \cite{uma2021learning} for a survey --- and we believe our task is in this category. To confirm this, we further annotated examples from Table~\ref{table:dataAmbiguity} as to their ambiguity (whether multiple choices seemed valid; see Table~\ref{table:dataAmbiguity}, ambiguous column). We find that for all \cdc s, more than 70\% of examples with disagreement are indeed ambiguous, explaining the relatively low values shown by traditional agreement measures that assume clear-cut decisions about assigning \cdc{} labels (Table~\ref{table:dataSpecificity}). The ambiguity problem is further exacerbated by the  multi-label nature of the task, increasing the number of possible \cdc{} label combinations and potential for disagreement.  
However, much recent work \cite{pavlick-kwiatkowski-2019-inherent,basile-etal-2021-need,leonardelli2021agreeing} shows it is possible (and also important) to design NLP models and evaluation measures that take task ambiguity into account. Consequently, we believe that \dsname{} will be valuable for future research.

To get a better perspective on comments to which the majority of annotators assigned the \textit{Other} label, 
an author manually inspected randomly selected 50 examples labeled with the \textit{Other} \cdc{} and 50 examples labeled with some other \cdc. Examples labeled as \textit{Other} were mainly  spam or non-offensive (mislabeled) comments. In contrast, different \cdc{} examples were mostly offensive, fitting well into one or more of the main six \cdc{} categories. The latter case accounts for the majority of examples.

\section{Results and Discussion}
\label{sec:result}

\subsection{Experimental Set-up}

\begin{table*}[ht]\centering
\resizebox{\textwidth}{!}{
\small
\begin{tabular}{l|cccccccc}
\hline
\cdc &  \multicolumn{2}{c}{\textbf{mBERT}} & \multicolumn{2}{c}{\textbf{XLM-RoBERTa}} & \multicolumn{2}{c}{\textbf{cseBERT}} & \multicolumn{2}{c}{\textbf{BERTi\'c}}   \\ 
Includes Explicit Expression & Yes & No & Yes & No & Yes & No & Yes & No  \\\hline \hline
Overall & 45.04 & 23.85 & 44.24 & 19.76 & \underline{56.70} & \textbf{28.46}  & \textbf{59.64} & \underline{26.53}   \\\hline 
Explicit  Expression & 60.12 & - & 61.65 & - & \underline{76.78} & - &\textbf{83.19} & - \\
Implicit  Expression & 21.54 & 21.76 & 16.49 & 16.53 & \textbf{26.86} & \textbf{26.45} & \textbf{26.86} & \underline{25.90} \\ \hline
Language Independent Disguise & 58.33 & 49.49 &59.80 & 50.51 & \underline{74.51} & \textbf{65.66} & \textbf{77.94} & \textbf{65.66} \\
World Knowledge-Based & 33.33 & 21.43 & 13.33 & ~~7.14 & \textbf{50.00} & \textbf{28.57} & \underline{46.67} & \underline{21.43} \\ \hline
Croatian Specific Disguise & 49.72 & 31.29 & 50.28 & 26.38 & \underline{64.92} & \underline{40.49} & \textbf{70.73} & \textbf{42.33} \\
Croatia Knowledge-Based & 40.91 & 23.29 & 38.81 & 15.75 & \underline{51.40} & \underline{23.29} & \textbf{55.59} & \textbf{27.40}  \\ \hline
Others & \textbf{11.93} & \textbf{11.43} & ~~8.52 & ~~8.00 & \underline{11.36} & \underline{10.86} & ~~6.25 & ~~5.71 \\ \hline 
\end{tabular}
}
\caption{Accuracy of the abusive comment on different \cdc s. $A_1$/$A_2$ where $A_1$ is accuracy on the unmodified test set and $A_2$ after removing \textit{Explicit Expression} examples. The best model is \textbf{bold} and second best \underline{underlined}.}
\label{table:accType} 
\end{table*}

For binary classification (i.e., \textit{Abuse} vs. \textit{Non-abuse}), we used the dataset from \citet{pelicon2021investigating}. We removed comments blocked for spam, deception \& trolling 
and use of a language other than Croatian, giving 4750/518/580 data points for training/validation/testing, respectively. We used four transformer-based models; two pre-trained on 100+ languages, namely mBERT~\citep{mbert} and XLM-RoBERTa base~\citep{conneau2020unsupervised} and two pre-trained on Croatian and 2-3 similar languages, namely cseBERT~\citep{ulcar-robnik2020finest} and BERTi\'c~\citep{ljubesic-lauc-2021-bertic}. We fine-tuned all models for the binary comment moderation task using default hyper-parameters for ten epochs, and selected the best model  based on validation F1 score.\footnote{On the corresponding test set, our model achieved macro F1 scores of 75.14, 76.72, 79.82, and 80.97 for mBERT, XLM-RoBERTa, cseBERT, and BERTi\'c, respectively, which is similar to previously reported results~\citep{pelicon2021investigating}.}

\begin{table*}[ht]
\centering
\small
\begin{tabular}{|l|r|r|r|r||r|r|r|r|}
\hline

\multirow{2}{*}{ \cdc} & \multicolumn{4}{c||}{\textbf{XLM-RoBERTa}} &  \multicolumn{4}{c|}{\textbf{BERTi\'c}}   \\
\cline{2-9}
  & 0 & 1 & 2 & 3 & 0 & 1 & 2 & 3 \\ \hline 
Explicit  Expression & 13.64 & 27.06 & 47.93 & 69.70  & 16.60 & 38.35 & 71.07 & 90.30  \\ 
Implicit Expression  &  57.75 & 31.75 & 18.91 & 9.90 & 74.79 & 46.74 & 29.45 & 19.80\\ \hline
Language Independent Disguise & 42.09 & 54.74 & 58.97 & 60.32 & 57.24 & 69.47 & 83.33 & 74.60 \\
World Knowledge-Based & 38.73 & 49.56 & 47.34 & 56.25 & 51.94 & 64.62 & 69.47 & 73.30 \\ \hline
Croatian Specific Disguise & 45.13 & 44.55 & 39.90 & 36.56 &  60.54 & 58.65 & 52.33 & 62.37 \\
Croatia Knowledge-Based  & 44.78 & 40.00 & 15.38 & 0.00 & 59.91 & 56.36 & 42.31 & 75.00\\ \hline
Others &50.21 & 19.70 & 9.71 & 6.85 &69.40 & 15.15 & 4.85 & 8.22 \\ \hline
\end{tabular}
 \caption{Performance of XLM-RoBERTa \& BERTi\'c based models per \cdc{} based on number of annotator's votes.}
\label{table:resultsCompare}
\end{table*}

\subsection{Quantitative Results}
\label{sec:quant}

Our primary goal is to study how models perform on fine-grained \cdc s, and we report accuracy on \dsname{} in Table~\ref{table:accType}. This number represents the proportion of comments from \dsname{ }that that a classification model (\textit{Abuse} vs. \textit{Non-abuse}) classified as \textit{Abuse} (by construction, all examples in \dsname{} should belong to \textit{Abuse}).
We present the overall accuracy of each annotated \cdc{} with and without the \textit{Explicit expression} \cdc. There are multiple insights from the results. For all \cdc s except \textit{Other}, cseBERT and BERTi\'c perform best. We confirm this using a permutation test~\citep{nichols2002nonparametric}: for all \cdc s except \textit{Other} the differences between the better of cseBERT/BERTić and the better of mBERT/XLM-RoBERTa, are statistically significant ($p \leq 0.05$). 
This again shows that a small multilingual Masked Language Model (MLM) with similar languages beats a massively multilingual MLM, similar to \citet{pelicon2021investigating}.

Among all the \cdc s, all models can easily identify the \textit{Explicit Expression} examples. Comparatively, \textit{Implicit Expression} is one of the most challenging \cdc, with more than 40\% difference between it and \textit{Explicit Expression}. This shows that it is hard for any model to identify implicit expression. At the same time, the \textit{Language Independent Disguise} \cdc{} is easier for models than the \textit{Croatian Specific Disguise} \cdc, with more than 7\% difference in the performance. On the \textit{Croatian Knowledge-Based} comments, cseBERT and BERTi\'c outperform mBERT and XLM-RoBERTa by a minimum 11\%. This, again, indicates that smaller multilingual MLM has comparatively more cultural information encoded.  

To better understand the effect of the \textit{Explicit Expression} comments, we removed all data points assigned the \textit{Explicit} \cdc{} label; results in Table~\ref{table:accType}. Overall performance drops  by $\geq$22\%, with a larger drop for cseBERT and BERTi\'c ($\geq$28\%). For both \textit{Local Context} \cdc s, there is a larger drop in performance ($\geq$26\%). This suggests we must find a better way to incorporate cultural knowledge into models. Furthermore, in Table~\ref{table:resultsCompare} we report the performance based on the number of annotator's votes, and show that our main observations still hold and are even more pronounced when considering data with high agreement.

\subsection{Qualitative Results}
\label{sec:quali}
Manual inspection of errors reveals some interesting patterns. 
Cases where all models fail almost always contain two or more \cdc s simultaneously, e.g.,
\example{Severaca moze glumiti jedino na camcu} \gloss{The only place where Severaca can act is a boat.} -- deliberate misspelling, reference to famous person, reference to local event).\footnote{Severaca refers to Severina, a regionally famous singer who was in a leaked explicit video taking place on a boat. The comment implies her acting skills are limited to pornography.} Moreover, examples where cseBERT and BERTić outperform mBERT and XLM-RoBERTa mostly require local context: e.g., \example{Opet. Retardesničaru.} \gloss{Again. You retarded right-wing extremist.} --  specific local word, wordplay only possible in Croatian). Finally, we find that examples on which all models perform well mostly contain explicit abuse with no misspelling, %
e.g., \example{Retard} \gloss{Retard}, 
which is in line with our empirical results.

\section{Conclusion}
\label{sec:con}

We present the \emph{\dslongname} (\dsname), a dataset annotated with context dependency categories (\cdc) of problematic examples for Croatian comment moderation. We annotated 2,240 blocked comments for Explicitness, Implicitness, Language Independent Disguise, World Knowledge-Based, Croatian Specific Disguise, and Croatia Knowledge-Based. We found that only 58.44\% had explicit expressions of abuse. This indicates that almost half the remaining examples are challenging (Croatian Specific Disguise alone accounting for $\approx 24\%$). This shows that addressing these categories of examples is very practically relevant. We tested four transformer-based models and found that explicit comments are the easiest and local context ones are hardest. We also found that language-specific multi-lingual language models better identify Croatian-specific blocked comments. Finally, we believe that \dsname{} will help design better models for Croatian comment moderation,  build a foundation for creating similar datasets in other languages, and develop novel methods by incorporating local context.

\section*{Ethical Consideration}
Our proposed dataset and models are to support more accurate and robust detection of online abuse. We anticipate that the high-quality and fine-grained \cdc{} labels in the dataset will advance research on online hate for low-resource languages. The dataset and models we present could, in principle, be used to train a generative hate speech model, but this is already possible using much larger datasets. Alternatively, the dataset and models could be used to understand current detection tools' limitations better and then attack them. However, we believe malicious actors are already manually employing similar attack methods to bypass the content rules of different platforms. Therefore, we believe that it is essential to understand how to attack the models and that our dataset will help the community fight such behavior by creating a more diverse dataset that leads to more robust models.

\section*{Acknowledgements}


We thank anonymous reviewers for their valuable feedback, and acknowledge financial support from several sources: the Slovenian Research Agency  via research core funding for the programme Knowledge Technologies (P2-0103) and the project Sovrag (Hate speech in contemporary conceptualizations of nationalism, racism, gender and migration, J5-3102); the UK EPSRC via the project Sodestream (Streamlining Social Decision Making for Improved Internet Standards, EP/S033564/1); and 
via the project RobaCOFI (Robust and adaptable comment filtering), which indirectly received funding from the European Union's Horizon 2020 research and innovation action programme via the AI4Media Open Call \#1 issued and executed under the AI4Media project (Grant Agreement no.\ 951911).
This paper reflects only the authors' views; the EC and the AI4Media project are not responsible for any use that may be made of the information contained herein.


\bibliography{ref,anthology}

\begin{thebibliography}{27}
\expandafter\ifx\csname natexlab\endcsname\relax\def\natexlab#1{#1}\fi

\bibitem[{Banko et~al.(2020)Banko, MacKeen, and Ray}]{banko2020unified}
Michele Banko, Brendon MacKeen, and Laurie Ray. 2020.
\newblock A unified taxonomy of harmful content.
\newblock In \emph{Proceedings of the fourth workshop on online abuse and
  harms}, pages 125--137.

\bibitem[{Basile et~al.(2021)Basile, Fell, Fornaciari, Hovy, Paun, Plank,
  Poesio, and Uma}]{basile-etal-2021-need}
Valerio Basile, Michael Fell, Tommaso Fornaciari, Dirk Hovy, Silviu Paun,
  Barbara Plank, Massimo Poesio, and Alexandra Uma. 2021.
\newblock \href {https://doi.org/10.18653/v1/2021.bppf-1.3} {We need to
  consider disagreement in evaluation}.
\newblock In \emph{Proceedings of the 1st Workshop on Benchmarking: Past,
  Present and Future}, pages 15--21, Online. Association for Computational
  Linguistics.

\bibitem[{Conneau et~al.(2020)Conneau, Khandelwal, Goyal, Chaudhary, Wenzek,
  Guzm{\'a}n, Grave, Ott, Zettlemoyer, and Stoyanov}]{conneau2020unsupervised}
Alexis Conneau, Kartikay Khandelwal, Naman Goyal, Vishrav Chaudhary, Guillaume
  Wenzek, Francisco Guzm{\'a}n, {\'E}douard Grave, Myle Ott, Luke Zettlemoyer,
  and Veselin Stoyanov. 2020.
\newblock Unsupervised cross-lingual representation learning at scale.
\newblock In \emph{Proceedings of the 58th Annual Meeting of the Association
  for Computational Linguistics}, pages 8440--8451.

\bibitem[{Devlin et~al.(2019)Devlin, Chang, Lee, and Toutanova}]{mbert}
Jacob Devlin, Ming-Wei Chang, Kenton Lee, and Kristina Toutanova. 2019.
\newblock Bert: Pre-training of deep bidirectional transformers for language
  understanding.
\newblock In \emph{Proceedings of the 2019 Conference of the North American
  Chapter of the Association for Computational Linguistics: Human Language
  Technologies, Volume 1 (Long and Short Papers)}, pages 4171--4186.

\bibitem[{Fleiss(1971)}]{fleiss1971measuring}
Joseph~L Fleiss. 1971.
\newblock Measuring nominal scale agreement among many raters.
\newblock \emph{Psychological bulletin}, 76(5):378.

\bibitem[{Fortuna et~al.(2019)Fortuna, da~Silva, Soler-Company, Wanner, and
  Nunes}]{Fortuna2019AHP}
Paula Fortuna, Jo{\~a}o~Rocha da~Silva, Juan Soler-Company, L.~Wanner, and
  S{\'e}rgio Nunes. 2019.
\newblock A hierarchically-labeled portuguese hate speech dataset.
\newblock \emph{Proceedings of the Third Workshop on Abusive Language Online}.

\bibitem[{Kurrek et~al.(2020)Kurrek, Saleem, and Ruths}]{kurrek2020towards}
Jana Kurrek, Haji~Mohammad Saleem, and Derek Ruths. 2020.
\newblock Towards a comprehensive taxonomy and large-scale annotated corpus for
  online slur usage.
\newblock In \emph{Proceedings of the Fourth Workshop on Online Abuse and
  Harms}, pages 138--149.

\bibitem[{Leonardelli et~al.(2021)Leonardelli, Menini, Aprosio, Guerini, and
  Tonelli}]{leonardelli2021agreeing}
Elisa Leonardelli, Stefano Menini, Alessio~Palmero Aprosio, Marco Guerini, and
  Sara Tonelli. 2021.
\newblock Agreeing to disagree: Annotating offensive language datasets with
  annotators’ disagreement.
\newblock In \emph{Proceedings of the 2021 Conference on Empirical Methods in
  Natural Language Processing}, pages 10528--10539.

\bibitem[{Ljube{\v{s}}i{\'c} and Lauc(2021)}]{ljubesic-lauc-2021-bertic}
Nikola Ljube{\v{s}}i{\'c} and Davor Lauc. 2021.
\newblock \href {https://www.aclweb.org/anthology/2021.bsnlp-1.5} {{BERT}i{\'c}
  - the transformer language model for {B}osnian, {C}roatian, {M}ontenegrin and
  {S}erbian}.
\newblock In \emph{Proceedings of the 8th Workshop on Balto-Slavic Natural
  Language Processing}, pages 37--42, Kiyv, Ukraine. Association for
  Computational Linguistics.

\bibitem[{Maronikolakis et~al.(2022)Maronikolakis, Wisiorek, Nann, Jabbar,
  Udupa, and Schuetze}]{maronikolakis-etal-2022-listening}
Antonis Maronikolakis, Axel Wisiorek, Leah Nann, Haris Jabbar, Sahana Udupa,
  and Hinrich Schuetze. 2022.
\newblock \href {https://doi.org/10.18653/v1/2022.findings-acl.87} {Listening
  to affected communities to define extreme speech: Dataset and experiments}.
\newblock In \emph{Findings of the Association for Computational Linguistics:
  ACL 2022}, pages 1089--1104, Dublin, Ireland. Association for Computational
  Linguistics.

\bibitem[{Mollas et~al.(2022)Mollas, Chrysopoulou, Karlos, and
  Tsoumakas}]{Mollas2022ETHOSAM}
Ioannis Mollas, Zoe Chrysopoulou, Stamatis Karlos, and Grigorios Tsoumakas.
  2022.
\newblock Ethos: a multi-label hate speech detection dataset.
\newblock \emph{Complex \& Intelligent Systems}, pages 1--16.

\bibitem[{Nichols and Holmes(2002)}]{nichols2002nonparametric}
Thomas~E Nichols and Andrew~P Holmes. 2002.
\newblock Nonparametric permutation tests for functional neuroimaging: a primer
  with examples.
\newblock \emph{Human brain mapping}, 15(1):1--25.

\bibitem[{Pavlick and Kwiatkowski(2019)}]{pavlick-kwiatkowski-2019-inherent}
Ellie Pavlick and Tom Kwiatkowski. 2019.
\newblock \href {https://doi.org/10.1162/tacl_a_00293} {Inherent disagreements
  in human textual inferences}.
\newblock \emph{Transactions of the Association for Computational Linguistics},
  7:677--694.

\bibitem[{Pelicon et~al.(2021{\natexlab{a}})Pelicon, Shekhar, Martinc,
  {\v{S}}krlj, Purver, and Pollak}]{pelicon-etal-2021-zero}
Andra{\v{z}} Pelicon, Ravi Shekhar, Matej Martinc, Bla{\v{z}} {\v{S}}krlj,
  Matthew Purver, and Senja Pollak. 2021{\natexlab{a}}.
\newblock \href {https://aclanthology.org/2021.hackashop-1.5} {Zero-shot
  cross-lingual content filtering: Offensive language and hate speech
  detection}.
\newblock In \emph{Proceedings of the EACL Hackashop on News Media Content
  Analysis and Automated Report Generation}, pages 30--34, Online. Association
  for Computational Linguistics.

\bibitem[{Pelicon et~al.(2021{\natexlab{b}})Pelicon, Shekhar, {\v{S}}krlj,
  Purver, and Pollak}]{pelicon2021investigating}
Andra{\v{z}} Pelicon, Ravi Shekhar, Bla{\v{z}} {\v{S}}krlj, Matthew Purver, and
  Senja Pollak. 2021{\natexlab{b}}.
\newblock Investigating cross-lingual training for offensive language
  detection.
\newblock \emph{PeerJ Computer Science}, 7:e559.

\bibitem[{Poletto et~al.(2021)Poletto, Basile, Sanguinetti, Bosco, and
  Patti}]{poletto2021resources}
Fabio Poletto, Valerio Basile, Manuela Sanguinetti, Cristina Bosco, and Viviana
  Patti. 2021.
\newblock Resources and benchmark corpora for hate speech detection: a
  systematic review.
\newblock \emph{Language Resources and Evaluation}, 55(2):477--523.

\bibitem[{Pollak et~al.(2021)Pollak, Robnik-{\v{S}}ikonja, Purver, Boggia,
  Shekhar, Pranji{\'c}, Salmela, Krustok, Paju, Linden, Lepp{\"a}nen, Zosa,
  Ul{\v{c}}ar, Freienthal, Traat, Cabrera-Diego, Martinc, Lavra{\v{c}},
  {\v{S}}krlj, {\v{Z}}nidar{\v{s}}i{\v{c}}, Pelicon, Koloski, Podpe{\v{c}}an,
  Kranjc, Sheehan, Boros, Moreno, Doucet, and
  Toivonen}]{pollak-etal-2021-embeddia}
Senja Pollak, Marko Robnik-{\v{S}}ikonja, Matthew Purver, Michele Boggia, Ravi
  Shekhar, Marko Pranji{\'c}, Salla Salmela, Ivar Krustok, Tarmo Paju,
  Carl-Gustav Linden, Leo Lepp{\"a}nen, Elaine Zosa, Matej Ul{\v{c}}ar, Linda
  Freienthal, Silver Traat, Luis~Adri{\'a}n Cabrera-Diego, Matej Martinc, Nada
  Lavra{\v{c}}, Bla{\v{z}} {\v{S}}krlj, Martin {\v{Z}}nidar{\v{s}}i{\v{c}},
  Andra{\v{z}} Pelicon, Boshko Koloski, Vid Podpe{\v{c}}an, Janez Kranjc, Shane
  Sheehan, Emanuela Boros, Jose~G. Moreno, Antoine Doucet, and Hannu Toivonen.
  2021.
\newblock \href {https://aclanthology.org/2021.hackashop-1.14} {{EMBEDDIA}
  tools, datasets and challenges: Resources and hackathon contributions}.
\newblock In \emph{Proceedings of the EACL Hackashop on News Media Content
  Analysis and Automated Report Generation}, pages 99--109, Online. Association
  for Computational Linguistics.

\bibitem[{R{\"o}ttger et~al.(2022)R{\"o}ttger, Seelawi, Nozza, Talat, and
  Vidgen}]{rottger-etal-2022-multilingual}
Paul R{\"o}ttger, Haitham Seelawi, Debora Nozza, Zeerak Talat, and Bertie
  Vidgen. 2022.
\newblock \href {https://aclanthology.org/2022.woah-1.15} {Multilingual
  {H}ate{C}heck: Functional tests for multilingual hate speech detection
  models}.
\newblock In \emph{Proceedings of the Sixth Workshop on Online Abuse and Harms
  (WOAH)}, pages 154--169, Seattle, Washington (Hybrid). Association for
  Computational Linguistics.

\bibitem[{R{\"o}ttger et~al.(2021)R{\"o}ttger, Vidgen, Nguyen, Waseem,
  Margetts, and Pierrehumbert}]{Rttger2021HateCheckFT}
Paul R{\"o}ttger, Bertie Vidgen, Dong Nguyen, Zeerak Waseem, Helen Margetts,
  and Janet Pierrehumbert. 2021.
\newblock Hatecheck: Functional tests for hate speech detection models.
\newblock In \emph{Proceedings of the 59th Annual Meeting of the Association
  for Computational Linguistics and the 11th International Joint Conference on
  Natural Language Processing (Volume 1: Long Papers)}, pages 41--58.

\bibitem[{Shekhar et~al.(2020)Shekhar, Pranjić, Pollak, Pelicon, and
  Purver}]{shekhar-etal-2020-automating}
Ravi Shekhar, Marko Pranjić, Senja Pollak, Andraž Pelicon, and Matthew
  Purver. 2020.
\newblock Automating news comment moderation with limited resources:
  Benchmarking in croatian and estonian.
\newblock \emph{Journal for Language Technology and Computational Linguistics
  (JLCL)}, 34(1).

\bibitem[{Ulčar and Robnik-Šikonja(2020)}]{ulcar-robnik2020finest}
M.~Ulčar and M.~Robnik-Šikonja. 2020.
\newblock \href {https://doi.org/10.1007/978-3-030-58323-1_11} {{FinEst BERT}
  and {CroSloEngual BERT}: less is more in multilingual models}.
\newblock In \emph{Text, Speech, and Dialogue {TSD 2020}}, volume 12284 of
  \emph{Lecture Notes in Computer Science}. Springer.

\bibitem[{Uma et~al.(2021)Uma, Fornaciari, Hovy, Paun, Plank, and
  Poesio}]{uma2021learning}
Alexandra~N Uma, Tommaso Fornaciari, Dirk Hovy, Silviu Paun, Barbara Plank, and
  Massimo Poesio. 2021.
\newblock Learning from disagreement: A survey.
\newblock \emph{Journal of Artificial Intelligence Research}, 72:1385--1470.

\bibitem[{Vidgen and Derczynski(2020)}]{vidgen2020directions}
Bertie Vidgen and Leon Derczynski. 2020.
\newblock Directions in abusive language training data, a systematic review:
  Garbage in, garbage out.
\newblock \emph{Plos one}, 15(12):e0243300.

\bibitem[{Waseem et~al.(2017)Waseem, Davidson, Ithica, Warmsley, and
  Weber}]{waseem2017understanding}
Zeerak Waseem, Thomas Davidson, NY~Ithica, Dana Warmsley, and Ingmar Weber.
  2017.
\newblock Understanding abuse: A typology of abusive language detection
  subtasks.
\newblock \emph{ACL 2017}, page~78.

\bibitem[{Wiegand et~al.(2021)Wiegand, Ruppenhofer, and
  Eder}]{wiegand2021implicitly}
Michael Wiegand, Josef Ruppenhofer, and Elisabeth Eder. 2021.
\newblock Implicitly abusive language--what does it actually look like and why
  are we not getting there?
\newblock In \emph{Proceedings of the 2021 Conference of the North American
  Chapter of the Association for Computational Linguistics: Human Language
  Technologies}, pages 576--587. Association for Computational Linguistics.

\bibitem[{Wiegand et~al.(2019)Wiegand, Ruppenhofer, and
  Kleinbauer}]{wiegand2019detection}
Michael Wiegand, Josef Ruppenhofer, and Thomas Kleinbauer. 2019.
\newblock Detection of abusive language: the problem of biased datasets.
\newblock In \emph{Proceedings of the 2019 conference of the North American
  Chapter of the Association for Computational Linguistics: human language
  technologies, volume 1 (long and short papers)}, pages 602--608.

\bibitem[{Zampieri et~al.(2019)Zampieri, Malmasi, Nakov, Rosenthal, Farra, and
  Kumar}]{zampieri2019predicting}
Marcos Zampieri, Shervin Malmasi, Preslav Nakov, Sara Rosenthal, Noura Farra,
  and Ritesh Kumar. 2019.
\newblock Predicting the type and target of offensive posts in social media.
\newblock In \emph{Proceedings of NAACL-HLT}, pages 1415--1420.

\end{thebibliography}
\bibliographystyle{acl_natbib}

\newpage
\section*{Appendix 1: Dataset Categories Examples}
\label{appx:ex}
In this section we provide some examples for the different categories. 

\paragraph{Explicit Expression} 
\begin{itemize}
    \item Use of Derogation (\textit{ti si nitko i ništa} -- \textit{You are a nobody.})
    \item Threatening Language (\textit{saznat ću gdje živiš} -- \textit{I will find out where you live.})
    \item Slur (\textit{retard} -- \textit{retard})
    \item Profanity (\textit{peder} -- \textit{fag})
\end{itemize}
\paragraph{Implicit Expression}: 
\begin{itemize}
    \item Abuse expressed using negated positive statements (\textit{“Gej je okej” je krivo} -- \textit{"Gay is ok" is wrong.})
    \item Abuse  phrased as a question (\textit{Zašto moramo tolerirati imigrante?} -- \textit{Why do we have to tolerate immigrants?})
    \item Abuse  phrased as an opinion (\textit{Staljin je imao pravi pristup.} -- \textit{Stalin had the right approach.})  
\end{itemize}

\paragraph{Language Independent Disguise}
\begin{itemize}
    \item Swaps of adjacent characters (\textit{jeib se -- f*ck you})
    \item Missing characters (\textit{jbi se})
    \item Missing word boundaries (\textit{jebise})
    \item Missing word boundaries (\textit{jebise})
    \item Added spaces between chars (\textit{j ebi se})
    \item Added spaces between chars (\textit{j ebi se})
    \item Added spaces between chars (\textit{j ebi se})
    \item substituting characters with “*”, “.” or similar. (\textit{je*i se})
    \item Leet speak spellings(\textit{j3b1 5e}).
\end{itemize}

\paragraph{World Knowledge-Based}
\begin{itemize}
    \item Momentary (knowledge of characters/events important at this point in time or for a relatively limited time) - e.g., \textit{Will Smith oscars slap}, \textit{Brexit}
    \item Long-term (more stable general knowledge) - e.g., \textit{The Pope}, \textit{Berlin wall}, \textit{The Beatles}
\end{itemize}

\paragraph{Croatian Specific Disguise}
\begin{itemize}

    \item ad-hoc constructed words that are understandable to locals (\textit{Svi prekodrinci su ološ} -- \textit{All X are scum.}, where X = \textit{prekodrinac}, an ad hoc invented word from \textit{preko ("across")} and \textit{Drina (name of a river)} denoting someone living across the Drina river -- i.e., Serbs), 
    \item misspelt important words in a way that is specific for croatian, mostly diacritics missing or wrong, like dj/dz for đ/dž, \textit{djubre/dubre} (instead of \textit{đubre} - piece of shit), \textit{cetnik} (instead of \textit{četnik} - member of a very unpopular military group), 
    \item using dialects (some abuse can sound very different in some dialects, containing words like \textit{Flundra, Droca, Štraca} -- easy woman), 
    \item idioms specific for Croatian (\textit{Promijenit ću ti lični opis.} -- \textit{I will change your personal description. i.e., I will break your face.}).
    \item other ways of using non-abusive words to create abusive context (which requires language knowledge to properly decypher) - sarcasm, substituting slurs for similarly sounding non-slurs, inventing abusive comparisons without abusive words on the spot -  e.g., \textit{bistar si ko mocvara} (\textit{your thinking is clear as a swamp}), \textit{u gnjurac} (\textit{gnjurac} is a bird, but sounds similar to \textit{kurac} -- \textit{d*ck}), referring to a person from the sea side as \textit{Tovar} (literal meaning is Donkey)
\end{itemize}

\paragraph{Croatia Knowledge-Based}
\begin{itemize}
    \item Momentary (knowledge of characters/events/facts important at this point in time or for a relatively limited time), e.g. \textit{Vili Beroš} (health minister during the Covid 19 pandemic), \textit{Uspinjača na sljeme} (controversial building project), 
    \item Long-term (more stable local knowledge) - e.g., \textit{HDZ} (a political party around for a long time), \textit{‘91} (year of the Croatian war of independence), \textit{Vrapče} (one of the most widely known Psychiatric institutions)
\end{itemize}

\section*{Appendix 2: Rule Description}
\label{appx:rule}
\begin{figure*}[!ht]
\begin{center}
\includegraphics[scale=0.40]{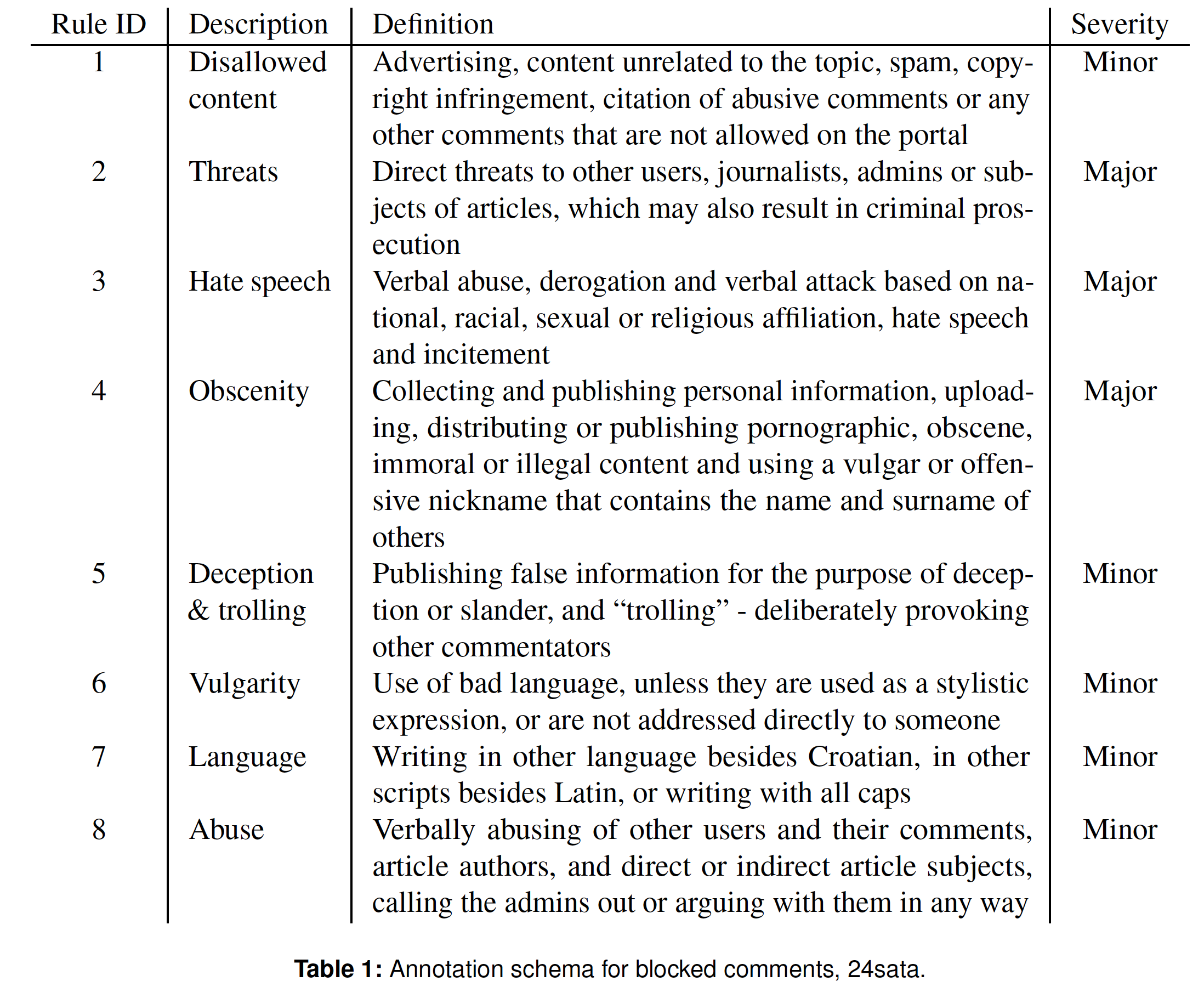} 
\caption{Rule description, reproduced from ~\citet{shekhar-etal-2020-automating}}
\label{fig:rule}
\end{center}
\end{figure*}

We have reproduced rule description from ~\citet{shekhar-etal-2020-automating} in Figure~\ref{fig:rule}.

\end{document}